\documentclass{article}

\PassOptionsToPackage{numbers, compress}{natbib}
\usepackage[final]{nips_2018}

\usepackage[utf8]{inputenc} % allow utf-8 input
\usepackage[T1]{fontenc}    % use 8-bit T1 fonts
\usepackage{hyperref}       % hyperlinks
\usepackage{url}            % simple URL typesetting
\usepackage{booktabs}       % professional-quality tables
\usepackage{amsfonts}       % blackboard math symbols
\usepackage{nicefrac}       % compact symbols for 1/2, etc.
\usepackage{microtype}      % microtypography

\usepackage{times}
\usepackage{epsfig}
\usepackage{graphicx}
\usepackage{graphics}
\usepackage{amsmath}
\usepackage{amssymb}
\usepackage{fontenc}
\usepackage{bm}
\usepackage{algorithm}
\usepackage{algorithmic}
\usepackage{multirow}
\usepackage{amsmath}
\usepackage{enumerate}
\usepackage{epstopdf}
\usepackage{setspace}
\usepackage{movie15}
\usepackage{subcaption}
\usepackage{color}
\usepackage{wrapfig}

\title{LMVP: Video Predictor with\\Leaked Motion Information}
\author{
  Dong Wang~$^1$, Yitong Li~$^1$, Wei Cao~$^2$, Liqun Chen~$^1$, Qi Wei~$^3$, Lawrence Carin~$^1$ \\
  $^1$~Duke University, $^2$~Tsinghua University, $^3$~JP Morgan \& Chase\\
  \texttt{\{dong.wang363, yitong.li, lcarin\}@duke.edu} \\
}

\begin{document}

\maketitle

\begin{abstract}
We propose a Leaked Motion Video Predictor (LMVP) to predict future frames by capturing the spatial and temporal dependencies from given inputs. The motion is modeled by a newly proposed component, motion guider, which plays the role of both learner and teacher. Specifically, it {\em learns} the temporal features from real data and {\em guides} the generator to predict future frames. The spatial consistency in video is modeled by an adaptive filtering network. To further ensure the spatio-temporal consistency of the prediction, a discriminator is also adopted to distinguish the real and generated frames. Further, the discriminator leaks information to the motion guider and the generator to help the learning of motion. The proposed LMVP can effectively learn the static and temporal features in videos without the need for human labeling. Experiments on synthetic and real data demonstrate that LMVP can yield state-of-the-art results.
\end{abstract}
% \vspace{-4mm}
% Supplemental: \url{http://s000.tinyupload.com/index.php?file_id=68257396191476330690}
\vspace{-4mm}
\section{Introduction}\label{sec:introduction}
Video combines structured spatial and temporal information in high dimensions. The strong spatio-temporal dependencies among consecutive frames in video greatly increases the difficulty of modeling. For instance, it is challenging to effectively separate moving objects from background, and predict a plausible future movement of the former~\cite{chao2017forecasting,henaff2017prediction,villegas2017decomposing,vondrick2017generating,walker2016uncertain,xue2016visual}. Though video is large in size and complex to model, video prediction is a task that can leverage the extensive online video data without the need of human labeling. Learning a good video predictor is an essential step toward understanding spatio-temporal modeling. These concepts can also be applied to various tasks, like weather forecasting, traffic-flow prediction, and disease control~\cite{wang2017deepsd,wang2018will,wang2016etcps}.

The recurrent neural network (RNN) is a widely used framework for spatio-temporal modeling. In most existing works, motion is estimated by the subtraction of two consecutive frames and the background is encoded by a convolutional neural network (CNN)~\cite{cai2017deep,patraucean2015spatio,villegas2017decomposing,villegas2017learning}. The CNN ensures spatial consistency, while temporal consistency is considered by the recurrent units, encouraging motion to smoothly progress through time. However, information in two consecutive frames is usually insufficient to learn the dynamics. Using 3D convolution to generate future frames can avoid these problems~\cite{vondrick2016generating,vondrick2017generating}, although generating videos by 3D-convolution usually lacks sharpness.

We propose a Leaked Motion Video Predictor (LMVP) for robust future-frame prediction. LMVP generates the prediction in an adversarial framework: we use a generative network to predict next video frames, and a discriminative network to judge the generated video clips.
For the motion part, we propose to learn the dynamics by introducing a motion guider, connecting the generator and the discriminator. The motion guider learns the motion feature through training on real video clips, and guides the prediction process by providing possible motion features. At the same time, in contrast with estimating motions by subtracting two consecutive frames, we allow the discriminator to leak high-level extracted dynamic features to the motion guider to further help the prediction. Such dynamic features provide more informative guidance about dynamics to the generator. The spatial dependencies of video are imposed by a convolutional filter network conditioned on the current frame. This idea is inspired by a conventional signal processing technique named adaptive filter, which can increase the flexibility of the neural network~\cite{jia2016dynamic}. It is assumed that, each pixel of the predicted frame is a nonlinear function of the neighborhood pixels of the current one, where the nonlinear function is implemented via LMVP as a deep neural network.

%In order to solve these two problems, we propose to learn the dynamics by introducing a motion guider in Section~\ref{subsec:motion_guide}, in an adversarial learning framework. The motion guider learns the motion feature through training with real video clips and guides the predicting process by providing possible motion features. The spatial dependencies of video are imposed by a convolutional filter network conditioned on the current frame. This idea is inspired by a conventional signal processing technique named adaptive filter, which can increase the flexibility of the neural network~\cite{jia2016dynamic}. Mathematically, each pixel of the predicted frame is a nonlinear function of the neighborhood pixels of the current one, where the nonlinear function is implemented via LMVP as a deep neural network. On top of the motion guider and the generator, a discriminator is introduced on the video clip to further improve the quality of the prediction~\cite{goodfellow2014generative}.
\vspace{-2mm}
\section{Models}\label{sec:models}
\begin{figure}[h]
	\centering
	\includegraphics[width=10cm]{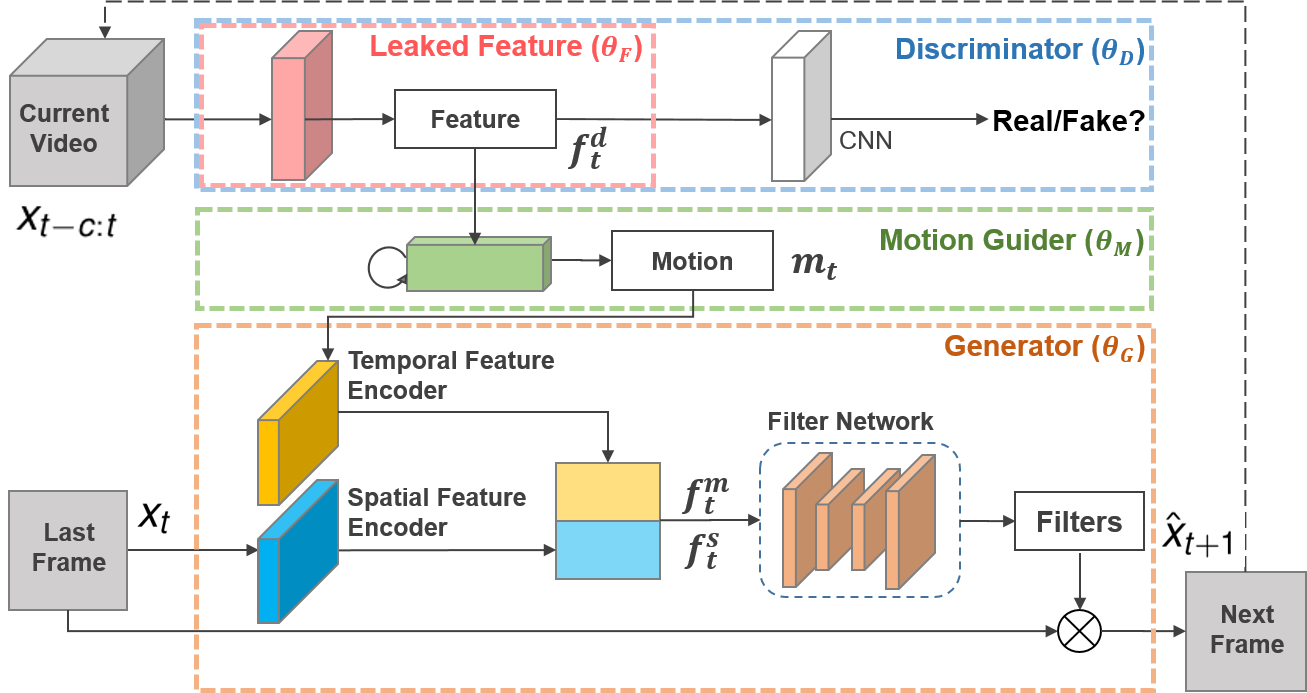}%framework.jpg%framework-new.jpg
	\caption{Model Framework.}\label{fig:framework}
\end{figure}

\vspace{-2mm}
The video frames are represented as $\bm x \in \mathbb{R}^{T \times H \times W \times C}$, where $T$ is the total number of frames, $H$ is the frame height, $W$ is the width and $C$ is the channel number. Given the first $T_0$ ($T_0 < T$) frames, the task is to predict the following $T-T_0$ frames. $\bm x_t$ and $\hat{\bm x}_t$ represent for real and predicted video frame at time $t$, respectively. The model framework is given in Figure~\ref{fig:framework}. It mainly contains a generator $G$, a motion guider $M$, and a discriminator $D$. $D$ distinguishes between the real and predicted video clips. $M$ learns the temporal dependencies among the video through the features leaked from $D$, and generator $G$ uses the output of motion guider $M$ to predict the next frame based on the current. 
\vspace{-5mm}
\subsection{Leaked Features from $D$ as Motion Signals}\label{subsec:discriminator}

The discriminator $D$ (shown in top of Figure~\ref{fig:framework}) is designed as both a discriminator and a motion feature extractor. The bottom layers of $D$ is a feature extractor $F(\cdot; \bm \theta_F)$, followed by several convolutional and fully connected layers to classify real/fake samples, parameterized by $\bm \theta_C$. Mathematically, given input video clips $\bm x_{t-c:t}$, we have $D(\bm x_{t-c:t}; \bm \theta_D) = \text{CNN}(F(\bm x_{t-c:t}; \bm \theta_F); \bm \theta_C)$,
where $\bm \theta_D = \{ \bm \theta_F, \bm \theta_C \}$. The extracted motion feature from $\bm x_{t-c:t}$ is denoted as $\bm f_t^d=F(\bm x_{t-c:t};\bm \theta_F)$, which is the input of motion guider $M$. 

The feature extractor $F$ is implemented as a convolutional network. The output $\bm f_t^d$ is expected to capture motion from $\bm x_{t-c:t}$. The difference between $\bm f_{t+1}^d$ and $\bm f_{t}^d$ is treated as the dynamic motion feature between two consecutive frames, which is denoted as $\bm m_t$. In contrast of the direct subtraction of two consecutive frames~\cite{villegas2017decomposing}, our dynamic motion feature is extracted from two consecutive video clips of length $c$. Since previous video frames are also included, it can still give reasonable output even if the model fails at previous time step. The discriminator loss can be written as
\begin{equation}\label{eq:discr_loss}
\mathcal{L}_{dis} (\bm \theta_D) = -\mathbb{E}_x[\log D(\bm x; \bm \theta_D)] - \mathbb{E}_{\hat{x}} [\log (1 - D(\hat{\bm x}; \bm \theta_D))].
\end{equation}
%A similar idea is also adopted in~\cite{villegas2017decomposing,villegas2017learning}. But their methods directly use raw pixel difference between two consecutive frames, while ours is from two video clips. 

\subsection{Learning and teaching game of $M$}\label{subsec:motion_guide}

To utilize the leaked motion information $\bm f_t^d$ from $D$, we introduce a motion guider module $M$, which is inspired by the leaky GAN model~\cite{guo2017long} for text generation task. 
The structure of $M$ is displayed in the green dotted box in Figure~\ref{fig:framework}.
$M$ has a recurrent structure that takes the extracted motion feature $\bm f_t^d$ as input at each time step $t$, and outputs a predicted motion feature $\hat{\bm m}_t$. 
%, which is then fed into the generator $G$.  
Specifically, the motion guider plays two roles in the model: learner and teacher. 

As a learner, $M$ learns the motion in video via leaked feature from $D$ from real video. At time step $t$, $M$ receives the leaked information $\bm f_t^d$ exacted by $D$, and predicts the dynamic motion feature between  time $t$ and $t+1$, by forcing $\hat{\bm m}_t = M(\bm f_t^d; \bm \theta_M)$ close to $\bm m_t = \bm f_{t+1}^d - \bm f_{t}^d$.
Denoting the parameters of the motion guider as $\bm \theta_M$, the learner loss function can be written as
\begin{equation}\label{eq:motion_guide_loss} %\bm f_{t+1}^d, \bm f_{t}^d
\mathcal{L}_{M}^L(\bm \theta_M) = \sum_t|| M(\bm f_t^d; \bm \theta_M) - (\bm f_{t+1}^d - \bm f_{t}^d )||_2^2.
\end{equation}
Note that only real video samples are used to update $\bm \theta_M$. The superscript $L$ means ``Leaner''.

As a teacher, $M$ serves as a guider by providing predicted dynamic motion features to $G$. 
During this step, $\bm \theta_M$ is fixed while the generator is updated under the guidance of $M$. Given the leaked features $\hat{\bm f}_t^d=F(\hat{\bm x}_{t-c:t};\theta_F)$ of predicted data $\hat{\bm x}_{t-c:t}$ at time $t$, the output $\hat{\bm m}_t = M(\hat{\bm f}_{t}^d; \bm \theta_M)$ serves as an input to the generator to predict the next frame $\hat{\bm x}_{t+1}$. This is detailed in the next section.

Since the dynamic motion feature is extracted from a real \emph{video clip} instead of a single frame, $M$ is robust against fail predictions at previous time step, i.e., even if the previous predicted frame diverges from the ground truth.
%\Qi{The separability of these two loss functions \eqref{eq:motion_guide_loss} and~\eqref{eq:motion_guide_fake} facilitates the possibility to update $\bm \theta_M$ and ($\bm \theta_D$, $\bm \theta_G$) alternatively}. For~\eqref{eq:motion_guide_loss}, $\bm \theta_M$ of the motion guider is updated using real data while the parameters in the generator and discriminator are kept fixed. For~\eqref{eq:motion_guide_fake}, the motion guider is fixed and $\hat{\bm m}_t$ is used to update the generator and discriminator. This reflects the learning and teaching process happened in motion guider. To summarize, the motion guider learns how to give real motion prediction in the learning step. In the teaching step, $\hat{\bm m}_t$ guides the generator and discriminator to give the next frame prediction. \Qi{In summary, this modular is referred to as motion guider} because it guides the prediction \Qi{using} the motion generation. \Qi{Note that} $\hat{\bm m}_t$ ensures motion consistency in predicted samples since $\hat{\bm f}_t^d$ does not directly depend on $\bm f_t^d$, i.e. the motion guider operates on the video sequence level. This helps to alleviate the accumulated error in standard recurrent networks. 

\subsection{Generating the next frame under guide from $M$}\label{subsec:predictor}
The structure of the generator is shown at the bottom of Figure~\ref{fig:framework}. It contains a spatial feature encoder, a temporal feature encoder, and a filter network. The spatial feature encoder is designed to learn the static background structure $\bm f_t^s$, while temporal feature $\bm f_t^m$ are computed from dynamic motion feature $\hat{\bm m}_t$. Note that only predicted samples have their motion guider output $\hat{\bm m}_t$ flow back to the generator. The spatio-temporal features $\bm f_t^s$ and $\bm f_t^m$ are concatenated and further fed into the filter network. The next frame is predicted by applying the generated adaptive filter from $[\bm f_t^s, \bm f_t^m]$ on the current frame. This technique is also known as visual transformation~\cite{vondrick2017generating}.

As mentioned in Section~\ref{subsec:motion_guide}, $M$ is updated during the learning step. When generating the next frame $\hat{\bm x}_{t+1}$ in the teaching step, $M$ is fixed and outputs the motion guide $\hat{\bm m}_t$. 
Specifically, to ensure $G$ generates the next frames following the guidance of $M$, the dynamic motion feature between the generated video clips at time $t+1$ and $t$, which is denoted as $\hat{\bm f}_{t+1}^d - \hat{\bm f}_t^d$, should be close to $\hat{\bm m}_t$ from $M$. Then, the generator is updated by minimizing the following loss function:
%Specifically, to ensure $G$ generates the next frames following the guidance of $M$, the feature $\hat{\bm f}_{t+1}^d$ of the generated frame $\hat{\bm x}_{t+1} = G(\hat{\bm x}_{t} ; \bm \theta_G)$ should be close to the feature taught by the guider $\hat{\bm m}_t + \hat{\bm f}^d_{t}$, the generator is updated by minimizing the following loss function:
\begin{equation}\label{eq:motion_guide_fake}
\mathcal{L}_{M}^T(\bm \theta_G) = \sum_t \|\left( F (\left[\hat{\bm x}_{t-c+1:t}, G(\hat{\bm x}_{t}, \hat{\bm m}_t ; \bm \theta_G)\right]) - \hat{\bm f}^d_{t} \right) - \hat{\bm m}_t  \|_2^2 ,
\end{equation}
where $\bm \theta_G$ includes all parameters in the generator $G$. The gradient is taken w.r.t $\bm \theta_G$, while $\hat{\bm f}_t^d$ and $\hat{\bm m}_t$ are treated as inputs. Note that $\hat{\bm x}_{t}$ and $\hat{\bm f}_t^d$ are the predicted output and leaked motion feature from time step $t$, respectively. The superscript $T$ in $\mathcal{L}_{M}^T$ indicates $M$ as a ``Teacher''.

The total loss function for the generator is 
\begin{equation}\label{eq:gen_loss}
\mathcal{L}_{gen}(\bm \theta_G) = \mathcal{L}_{recons}(\bm \theta_G) + \gamma \mathcal{L}_{M}^T(\bm \theta_G). %\gamma_1 \mathcal{L}_{adv}(\hat{\bm x}_{t-c+1:t+1}) + \gamma_2 
\end{equation}

$\mathcal{L}_{recons}$ is the reconstruction loss function of $\bm x_{t+1}, \hat{\bm x}_{t+1}$, including a pixel-wise cross-entropy/MSE loss and the gradient difference loss (GDL)~\cite{mathieu2015deep}. %defined as 
% $\mathcal{L}_{gdl} = \sum_{i,j} \left| |x_{ij} - x_{i(j-1)}| - |\hat{x}_{ij} - \hat{x}_{i(j-1)}|\right|^{\alpha} + \left| |x_{ij} - x_{(i-1)j}| - |\hat{x}_{ij} - \hat{x}_{(i-1)j}|\right|^{\alpha}$. $\alpha \geqslant 1$ is an integer, controlling the sharpness of the prediction. Note that GDL loss is designed to model the neighborhood pixel differences. 
%The third loss term $\mathcal{L}_{adv}(\hat{\bm x}_{t-c+1:t+1})$ comes from the discriminator, which is defined as $\mathcal{L}_{adv}(\hat{\bm x}_{t-c+1:t+1}) = -\mathbb{E}_{\hat{x}} [\log \mathcal D(\hat{\bm x}_{t-c+1:t+1}; \theta_D)]$. The generator loss~\eqref{eq:gen_loss} forces the $\hat{\bm x}_{t+1}$ to be as close to the ground truth $\bm x_{t+1}$ as possible. 
The whole model is updated iteratively for each component. A pseudo-algorithm is given in Alg.~\ref{alg:lmvp} in the Appendix. The discriminator loss~\eqref{eq:discr_loss} is first evaluated and $\bm \theta_D$ is updated. Then the motion guider parameters $\bm \theta_M$ are updated using only real samples of $\bm x$. The generated parameters $\bm \theta_G$ is updated using loss function~\eqref{eq:gen_loss}. % , together with the leaked feature extractor parameters $\bm \theta_F$ are updated using loss function~\eqref{eq:gen_loss}. 
In practice, Adam~\cite{kingma2014adam} is used to perform the gradient descent optimization.

\begin{wraptable}{r}{0.5\textwidth}
\vspace{-6mm}
	\centering
	\resizebox{.5\textwidth}{!}{
	\begin{tabular}{l|c|c|c}
		\hline
		& BCE & SSIM & PSNR  \\  
 		\hline
        ConvLSTM \cite{xingjian2015convolutional} & $8.96 \times 10^{-2}$ & $0.61$ & $10.74$ \\
        \hline
        FPM \cite{srivastava2015unsupervised} & $8.33 \times 10^{-2}$ & - &  - \\
        \hline
		DFN \cite{jia2016dynamic} & $6.89 \times 10^{-2}$ & $0.83$ & $18.4$ \\	
        \hline
		LMVP & \textbf{$6.13 \times 10^{-2}$} & $0.87$ & $19.6$\\
		\hline 
	\end{tabular}
	}
    \caption{Binary cross-entropy, SSIM and PSNR scores results on Moving MNIST dataset.}
	\label{tab:MNIST_result}
	\vspace{-2mm}
\end{wraptable}

\section{Experiments} \label{sec:experiments}

\textbf{Moving MNIST}: Each video in the Moving MNIST dataset~\cite{srivastava2015unsupervised} has $20$ frames in total, with two handwritten digits bouncing inside a $64 \times 64$ patch. Given $10$ frames, the task is to predict the motion of the digits of the following $10$ frames. We follow the same training and testing procedure as~\cite{srivastava2015unsupervised}. Evaluation metrics include Binary Cross Entropy (BCE), Peak Signal to Noise Ratio (PSNR), and Structural Similarity Index Measure (SSIM)~\cite{wang2004image} between the ground truth $\bm x$ and the prediction $\hat{\bm x}$. Small values of BCE or large values of SSIM and PSNR indicate good prediction results. In this task, we need to keep the digit shape the same across time (spatial consistency) while giving them reasonable movements (temporal consistency).

Table \ref{tab:MNIST_result} gives the comparison of LMVP and baseline models. The BCE of LMVP achieves $0.061$ per pixel over $10$ frames, which is better than state-of-the-art models~\cite{xingjian2015convolutional,srivastava2015unsupervised,jia2016dynamic,villegas2017decomposing}. The predictions from LMVP and DFN~\cite{jia2016dynamic} are shown in Figure \ref{fig:visual}. Input is given in the first row, followed by ground truth of the output, and results from DFN model and our LMVP model. To prove that our model has consistently good result in to the future, Figure \ref{fig:evaluations} in the Appendix gives the SSIM and PSNR comparison over $(t=T_0+1,\cdots,T)$. LMVP achieves higher SSIM and PSNR scores than other baseline models through all time steps.
\begin{figure}[htb]
    \centering
    \includegraphics[width=\textwidth]{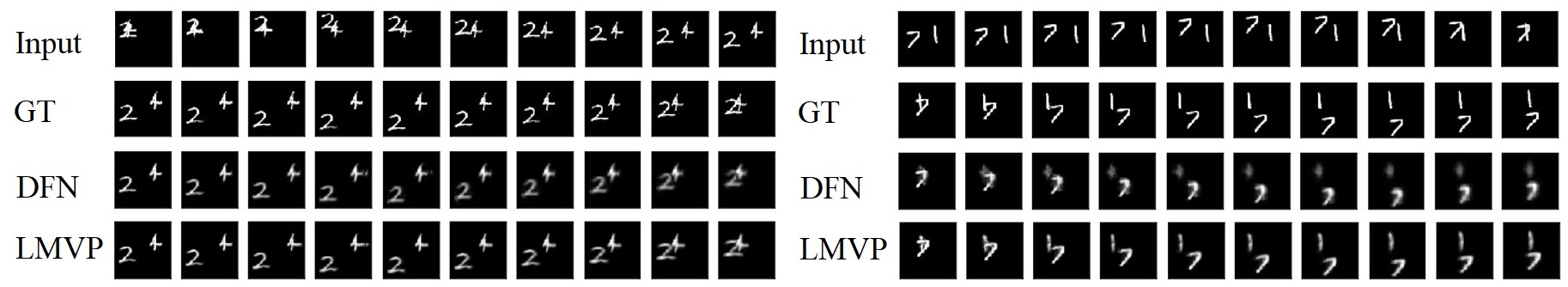}    
    \caption{Two prediction examples for the Moving MNIST dataset. From top to down: concatenation of input and ground truth, prediction result of DFN, and prediction result of our model.}
    \label{fig:visual}
\end{figure}
\begin{wraptable}{r}{0.5\textwidth}
	\centering
	\begin{tabular}{l|c|c|c}
		\hline
		& MSE & SSIM & PSNR  \\
		\hline
        {Last Frame} &  $10.34 \times 10^{-3}$ & $0.83$ & $22.11$ \\    
        \hline
		DFN \cite{jia2016dynamic} & $3.08 \times 10^{-3}$ & $0.92$ & $26.95$ \\	
		\hline
		LMVP & $2.67 \times 10^{-3}$ & $0.927$ & $27.23$ \\
		\hline 
	\end{tabular}
    \caption{MSE (per pixel), SSIM and PSNR scores results on highway driving video dataset.}
	\label{tab:highway}
	\vspace{-4mm}
\end{wraptable}
\textbf{Highway Drive}:
The dataset contains videos were collected from a car-mounted camera during car driving on a highway. The videos contain rich temporal dynamics, including both self-motion of the car and the motion of other objects in the scene~\cite{lotter2016deep}. Following the setting used in \cite{jia2016dynamic}, we split the approximately $20,000$ frames of the $30$-minutes video into a training set of $16,000$ frames and a test set of $4,000$ frames. Each frame is of size $64 \times 64$. The task is to predict three frames in the future given the past three. 

\begin{figure}[htb]
    \centering
    \includegraphics[width=0.7\textwidth]{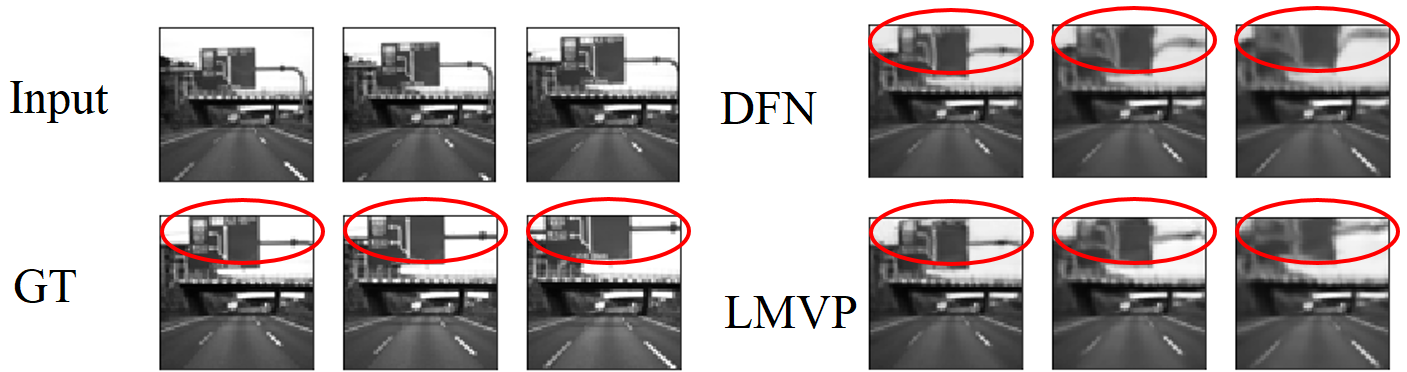}  
    \caption{Qualitative prediction examples for the highway driving video dataset. From top to down, left to right: concatenation of input and ground truth, prediction result of DFN, and prediction result of our LMVP.}
    \label{fig:highway_comp}
\end{figure}

The prediction results are compared in Table~\ref{tab:highway} and two samples from the test set are selected in Figure \ref{fig:highway_comp}. In the prediction results of DFN, the rail of the guidepost becomes curving. However, in the prediction results of LMVP, the rail keeps straight in the first and second predicted frames. To help the visual comparison, this part has been highlighted by a red circle.
% We have also compared \Qi{LMVP} with the state-of-the-art work~\cite{lotter2016deep}. To be consistent with the setting of \cite{lotter2016deep}, we \Qi{trained} our model to predict the next frame instead of \Qi{the following $3$ ones}, given the previous $3$ frames. We \Qi{obtained} a MSE loss of $1.39 \times 10^{-3}$ on the test set, while $1.16 \times 10^{-3}$ \Qi{was} achieved by the method in ~\cite{lotter2016deep}. \Qi{Considering that~\cite{lotter2016deep} is specially designed} for next-frame prediction, \Qi{the proposed model outperforms it with a more general setting}.
\vspace{-7mm}
\section{Conclusion}\label{sec:conclusion}
We have proposed the Leaked Motion Video Predictor (LMVP) to handle the spatio-temporal consistency in video prediction. For the dynamics in video, the motion guider learns motion features from real data and guides the prediction. Since the motion guider learns features from video sequences, it is more robust compared to using only single frames as input. For structures of the background, the adaptive filter generates input-aware filters when predicting the next frame, ensuring spatial consistency. Further, A discriminator is adopted to further improve the prediction result. On both synthetic and real datasets, LMVP shows superior results over the state-of-the-art approaches. 

\bibliographystyle{plainalt}
\bibliography{ref}

\begin{thebibliography}{10}

\bibitem{cai2017deep}
H.~Cai, C.~Bai, Y.-W. Tai, and C.-K. Tang.
\newblock Deep video generation, prediction and completion of human action
  sequences.
\newblock {\em arXiv preprint arXiv:1711.08682}, 2017.

\bibitem{chao2017forecasting}
Y.-W. Chao, J.~Yang, B.~Price, S.~Cohen, and J.~Deng.
\newblock Forecasting human dynamics from static images.
\newblock In {\em IEEE CVPR}, 2017.

\bibitem{guo2017long}
J.~Guo, S.~Lu, H.~Cai, W.~Zhang, Y.~Yu, and J.~Wang.
\newblock Long text generation via adversarial training with leaked
  information.
\newblock {\em AAAI}, 2018.

\bibitem{henaff2017prediction}
M.~Henaff, J.~Zhao, and Y.~LeCun.
\newblock Prediction under uncertainty with error-encoding networks.
\newblock {\em arXiv preprint arXiv:1711.04994}, 2017.

\bibitem{jia2016dynamic}
X.~Jia, B.~De~Brabandere, T.~Tuytelaars, and L.~V. Gool.
\newblock Dynamic filter networks.
\newblock In {\em NIPS}, 2016.

\bibitem{kingma2014adam}
D.~P. Kingma and J.~Ba.
\newblock Adam: A method for stochastic optimization.
\newblock {\em arXiv preprint arXiv:1412.6980}, 2014.

\bibitem{lotter2016deep}
W.~Lotter, G.~Kreiman, and D.~Cox.
\newblock Deep predictive coding networks for video prediction and unsupervised
  learning.
\newblock {\em ICLR}, 2017.

\bibitem{mathieu2015deep}
M.~Mathieu, C.~Couprie, and Y.~LeCun.
\newblock Deep multi-scale video prediction beyond mean square error.
\newblock {\em ICLR}, 2017.

\bibitem{patraucean2015spatio}
V.~Patraucean, A.~Handa, and R.~Cipolla.
\newblock Spatio-temporal video autoencoder with differentiable memory.
\newblock {\em ICLR}, 2016.

\bibitem{srivastava2015unsupervised}
N.~Srivastava, E.~Mansimov, and R.~Salakhudinov.
\newblock Unsupervised learning of video representations using lstms.
\newblock In {\em ICML}, 2015.

\bibitem{villegas2017decomposing}
R.~Villegas, J.~Yang, S.~Hong, X.~Lin, and H.~Lee.
\newblock Decomposing motion and content for natural video sequence prediction.
\newblock {\em ICLR}, 2017.

\bibitem{villegas2017learning}
R.~Villegas, J.~Yang, Y.~Zou, S.~Sohn, X.~Lin, and H.~Lee.
\newblock Learning to generate long-term future via hierarchical prediction.
\newblock {\em aICML}, 2017.

\bibitem{vondrick2016generating}
C.~Vondrick, H.~Pirsiavash, and A.~Torralba.
\newblock Generating videos with scene dynamics.
\newblock In {\em NIPS}, 2016.

\bibitem{vondrick2017generating}
C.~Vondrick and A.~Torralba.
\newblock Generating the future with adversarial transformers.
\newblock In {\em CVPR}, 2017.

\bibitem{walker2016uncertain}
J.~Walker, C.~Doersch, A.~Gupta, and M.~Hebert.
\newblock An uncertain future: Forecasting from static images using variational
  autoencoders.
\newblock In {\em ECCV}, 2016.

\bibitem{wang2017deepsd}
D.~Wang, W.~Cao, J.~Li, and J.~Ye.
\newblock Deepsd: supply-demand prediction for online car-hailing services
  using deep neural networks.
\newblock In {\em 2017 IEEE 33rd International Conference on Data Engineering
  (ICDE)}. IEEE, 2017.

\bibitem{wang2016etcps}
D.~Wang, W.~Cao, M.~Xu, and J.~Li.
\newblock Etcps: An effective and scalable traffic condition prediction system.
\newblock In {\em International Conference on Database Systems for Advanced
  Applications}. Springer, 2016.

\bibitem{wang2018will}
D.~Wang, J.~Zhang, W.~Cao, J.~Li, and Y.~Zheng.
\newblock
\newblock When will you arrive? estimating travel time based on deep neural
  networks. AAAI, 2018.

\bibitem{wang2004image}
Z.~Wang, A.~C. Bovik, H.~R. Sheikh, and E.~P. Simoncelli.
\newblock Image quality assessment: from error visibility to structural
  similarity.
\newblock {\em IEEE transactions on image processing}, 2004.

\bibitem{xingjian2015convolutional}
S.~Xingjian, Z.~Chen, H.~Wang, D.-Y. Yeung, W.-K. Wong, and W.-c. Woo.
\newblock Convolutional lstm network: A machine learning approach for
  precipitation nowcasting.
\newblock In {\em NIPS}, 2015.

\bibitem{xue2016visual}
T.~Xue, J.~Wu, K.~Bouman, and B.~Freeman.
\newblock Visual dynamics: Probabilistic future frame synthesis via cross
  convolutional networks.
\newblock In {\em NIPS}, 2016.

\end{thebibliography}

\clearpage

\appendix
\section{Model training}\label{supp:training}
The model is first pre-trained by iteratively updating the parameters of $D$ and $G$. In each iteration, we first update $\bm \theta_D$ by minimizing the loss $\mathcal{L}_{dis}$ in Equation~\eqref{eq:discr_loss}; then, $\bm \theta_F$, $\bm \theta_M$, and $\bm \theta_G$ are jointly updated by minimizing the loss $\mathcal{L}_{gen}$ in Equation~\eqref{eq:gen_loss} with $\gamma=0$. We found that the above pre-training technique can empirically stabilize the generation process and learn useful leaked information from discriminator.

In the main algorithm loop, $D$, $M$, and $G$ are trained iteratively. The algorithm outline is given in Algorithm~\ref{alg:lmvp}. Firstly, $\bm \theta_D$ is updated according to discriminator loss $\mathcal{L}_{dis}$ while $\bm \theta_M$ and $\bm \theta_G$ are kept fixed. Secondly, $\mathcal{L}_{M}(\bm \theta_M)$ defined in Equation~\eqref{eq:motion_guide_loss} is evaluated to update $\bm \theta_M$ while $\bm \theta_D$ and $\bm \theta_G$ remain unchanged. The third step is to update $\bm \theta_G$ by minimizing loss $\mathcal{L}_{gen}$ in Equation~\eqref{eq:gen_loss} with $\gamma>0$. 
Note that, in both pre-train and main algorithm loop, all the initial hidden states in recurrent architecture are set to zero. The gradient is updated by Adam~\cite{kingma2014adam}.

{
\begin{algorithm}[htb]
\caption{Leaked Motion Video Prediction}\label{alg:lmvp}
\begin{algorithmic}[1]
\STATE \textbf{Input}: Training videos $\bm V = \{x_{1:T}\}$.
\STATE \textbf{Output}: Parameters $\bm \theta_G$, $\bm \theta_D$, $\bm \theta_F$ and $\bm \theta_M$.
\\\hrulefill
\STATE Initialize the discriminator, generator, and motion guider with random weights. Initial hidden states in the model are set to zeros. 
\STATE Pre-train $\bm \theta_D$ using videos in training dataset as positive samples and output from generator as negative samples.
\STATE Pre-train $\bm \theta_G$, $\bm \theta_F$ and $\bm \theta_M$ using leaked features from $\bm \theta_F$ according to loss $\mathcal{L}_{recons}$.
\STATE Repeat the two pre-train steps iteratively until convergence.
\FOR{$iter = 1$ to $max\_iter$}
\STATE Sample a mini-batch of real video clips $\{ \bm x \}$ and generate fake video clips $\{ \hat{\bm x} \}$ according to the input.
\STATE $//$ \textbf{Train Discriminator}
  \STATE Fix $\bm \theta_M$ and $\bm \theta_G$, update discriminator parameters $\bm \theta_D$ by $\frac{\partial \mathcal{L}_{dis}}{\partial \bm \theta_D}$.  
  %\ENDFOR
\STATE $//$ \textbf{Train Motion Guider}  
%   \FOR{$i=1$ to $\text{iter}_M$}
	\STATE Compute $\bm f_t^d$ and $\bm m_t$ using real data % with $\bm \theta_D$ and $\bm \theta_M$, respectively.
  \STATE Compute motion guide (learner) loss $\mathcal{L}_{M}^L$ in Equation~\eqref{eq:motion_guide_loss} using $\bm m_t$ and $\bm f_t^d$ computed above. 
  \STATE Fix $\bm \theta_G$ and $\bm \theta_D$, update $\bm \theta_M$ by $\frac{\partial \mathcal{L}_{M}^L}{\partial \bm \theta_M}$.  
%   \ENDFOR
\STATE $//$ \textbf{Train generator} 
%    \FOR{$i=1$ to $\text{iter}_G$}
 \STATE Compute the prediction $\hat{\bm x}_{t+1} = G(\bm x_t, \bm m_t; \bm \theta_G)$ and loss $\mathcal{L}_{M}^T$ in~\eqref{eq:motion_guide_fake} using $\hat{\bm m}_t$ and $\hat{\bm f}_t^d$ from generated samples in Equation~\eqref{eq:gen_loss}.
 \STATE Fix $\bm \theta_M$ and $\bm \theta_D$, update $\bm \theta_G$ by $\frac{\partial \mathcal{L}_{gen}}{\partial \bm \theta_G}$.
% \STATE Fix $\bm \theta_M$, update $\bm \theta_P$ by $\frac{\partial \mathcal{L}_{\hat{M}}}{\partial \bm \theta_P}$ and update $\bm \theta_D$ by $\frac{\partial \mathcal{L}_{\hat{M}}}{\partial \bm \theta_D}$
%  \ENDFOR
\ENDFOR
\end{algorithmic}
\end{algorithm}
}

\newpage
\section{Experiment Result on Moving MNIST Dataset}\label{supp:moving_mnist}
{The} predictions generated by a LMVP model and a DFN model~\cite{jia2016dynamic} are {displayed} in Figure \ref{fig:visual}. Frames in the first line are input sequences and ground truth sequences.
Frames generated by a DFN model and our LMVP model are shown in the second and the third line, respectively. {Visually, the prediction of LMVP is better than DFN. This is further confirmed quantitatively in Table \ref{tab:psrn}.}

\begin{figure}[htb]
    \centering
    \includegraphics[width=\textwidth]{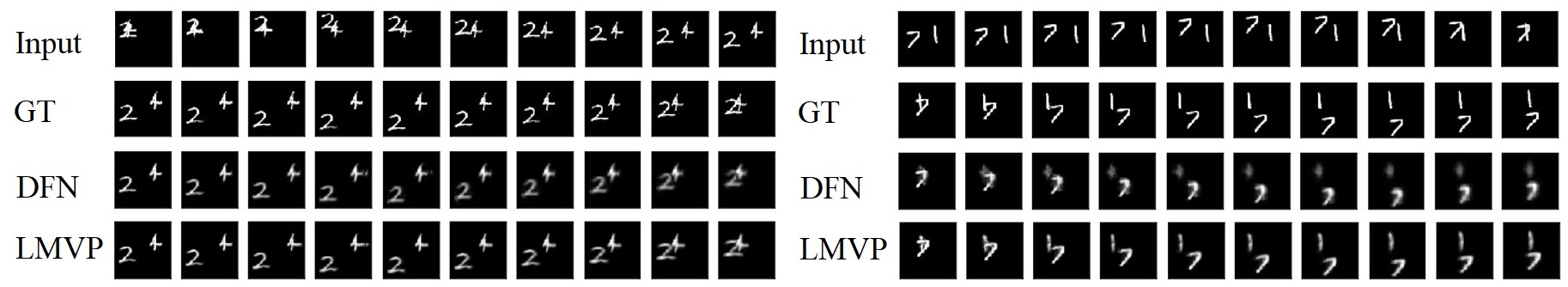}    
    \caption{Two prediction examples for the Moving MNIST dataset. From top to down: concatenation of input and ground truth, prediction result of DFN, and prediction result of our model.}
    \label{fig:visual}
\end{figure}

\begin{table}[h]
	\centering
	\begin{tabular}{l|c|c|c|c|c ||c|c|c|c|c }
		\hline
		PSNR & $2$ & $4$ & $6$ & $8$ & $10$ & $2$ & $4$ & $6$ & $8$ & $10$\\
		\hline
        DFN  & $23.42$ & $21.19$ & $19.01$ & $18.21$ & $18.46$ & $18.70$ & $18.38$ & $18.32$ & $17.91$ & $17.76$\\
        \hline
	    LMVP & $25.25$ & $22.99$ & $22.01$ & $20.28$ & $20.44$ & $21.73$ & $21.29$ & $21.05$ & $20.85$ & $20.45$ \\
%         \hline
% 		DFN  & $18.70$ & $18.38$ & $18.32$ & $17.91$ & $17.76$ \\
%         \hline
% 	    LMVP & $21.73$ & $21.29$ & $21.05$ & $20.85$ & $20.45$ \\
		\hline 
	\end{tabular}
    \caption{PSNR scores of prediction frames for each even time step shown in Figure~\ref{fig:visual} on Moving MNIST dataset.}
	\label{tab:psrn}
\end{table}

\begin{figure}[htb]

    \centering
    \begin{subfigure}[b]{0.35\textwidth}
        \includegraphics[width=\textwidth]{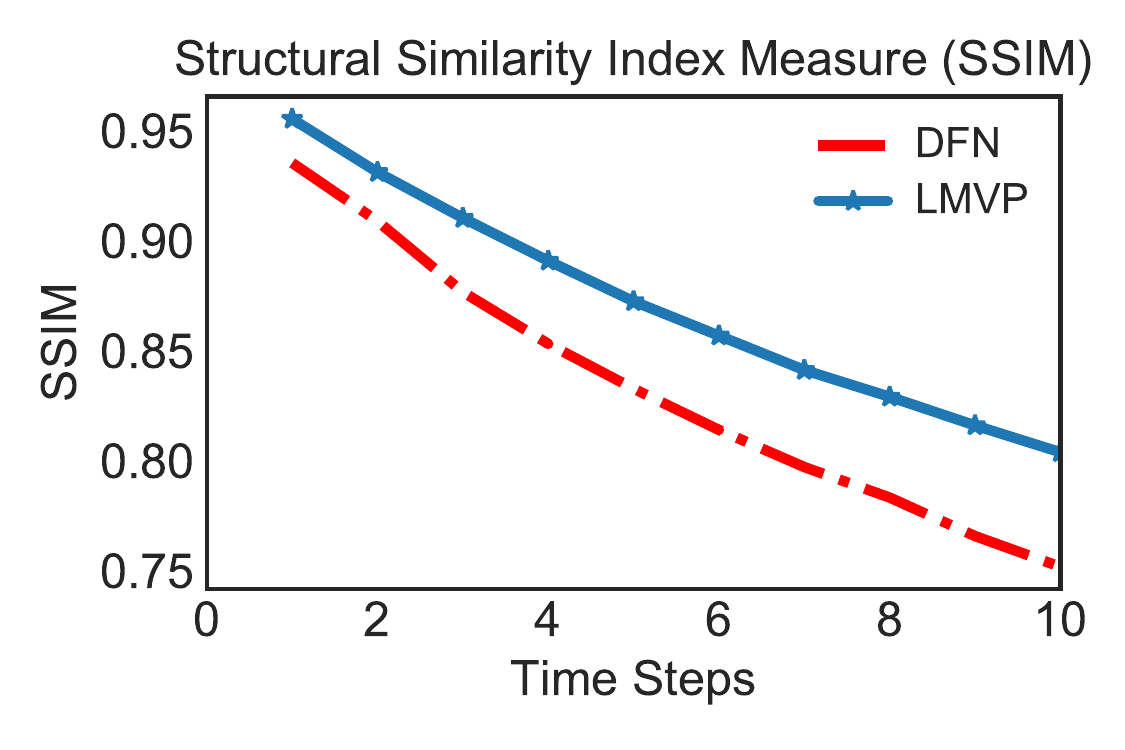}
        \label{fig:ssim}
    \end{subfigure}
    \begin{subfigure}[b]{0.35\textwidth}
        \includegraphics[width=\textwidth]{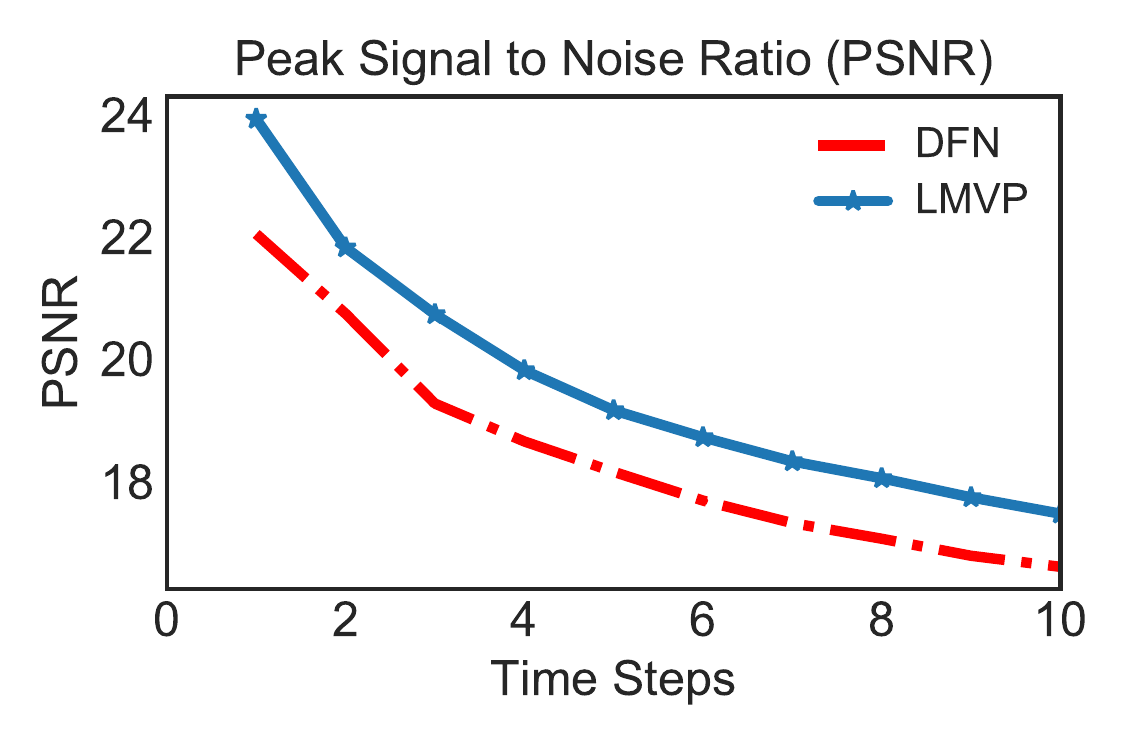}
        \label{fig:psnr}
    \end{subfigure}    
    \vspace{-7mm}
    \caption{Evaluation result of DFN and ours over different time step (from $1^{st}$ frame prediction scores to $10^{th}$ frame prediction scores). The proposed model gets higher SSIM and PSNR scores than baselines through all time steps.}
    \label{fig:evaluations}
\end{figure}

{To demonstrate it more clearly, Figure \ref{fig:evaluations} displays} the prediction evaluation of DFN and our model over different time step $(t=T_0+1,\cdots,T)$. LMVP {achieves} higher SSIM and PSNR scores than other baseline models through all time steps.

% Figure~\ref{fig:evaluations} gives the comparison on SSIM and PSNR with different number of predicted frames on mnist dataset. The proposed model consistently outperforms the baseline model.
\end{document}